\documentclass[journal]{IEEEtran}

\ifCLASSINFOpdf
\else
\fi
\usepackage{graphicx} 
\usepackage{amsmath}
\usepackage{algorithm}
\usepackage{algpseudocode}
\usepackage{algorithmicx}
\usepackage{float}
\usepackage{caption}
\usepackage{subfig}
\usepackage{color}
\usepackage{multirow}
\usepackage{cite}

\begin{document}

\title{Motif-topology improved Spiking Neural Network for the Cocktail Party Effect and McGurk Effect}
\author{Shuncheng~Jia, Tielin~Zhang, Ruichen~Zuo and Bo~Xu
\thanks{Shuncheng~Jia and Tielin Zhang are with Institute of Automation, Chinese Academy of Sciences (CASIA), Beijing 100190, China, and School of Artificial Intelligence, University of Chinese Academy of Sciences (UCAS), Beijing 100049, China.}
\thanks{Ruichen~Zuo is with School of Information and Electronics, Beijing Institute of Technology (BIT), Beijing 100081, China.}
\thanks{Bo Xu is with CASIA, UCAS, and Center for Excellence in Brain Science and Intelligence Technology, Chinese Academy of Sciences, Shanghai 200031, China.}
\thanks{Shuncheng Jia and Tielin Zhang are co-first authors of this paper. The corresponding authors are Tielin Zhang and Bo Xu, with emails of tielin.zhang@ia.ac.cn and xubo@ia.ac.cn.}
}

\markboth{Latex}%
{Jia \MakeLowercase{\textit{et al.}}}

\maketitle

\begin{abstract}
Network architectures and learning principles are playing key in forming complex functions in artificial neural networks (ANNs) and spiking neural networks (SNNs). SNNs are considered the new-generation artificial networks by incorporating more biological features than ANNs, including dynamic spiking neurons, functionally specified architectures, and efficient learning paradigms. Network architectures are also considered embodying the function of the network. Here, we propose a Motif-topology improved SNN (M-SNN) for the efficient multi-sensory integration and cognitive phenomenon simulations. The cognitive phenomenon simulation we simulated includes the cocktail party effect and McGurk effect, which are discussed by many researchers. Our M-SNN constituted by the meta operator called network motifs. The source of 3-node network motifs topology from artificial one pre-learned from the spatial or temporal dataset. In the single-sensory classification task, the results showed the accuracy of M-SNN using network motif topologies was higher than the pure feedforward network topology without using them. In the multi-sensory integration task, the performance of M-SNN using artificial network motif was better than the state-of-the-art SNN using BRP (biologically-plausible reward propagation). Furthermore, the M-SNN could better simulate the cocktail party effect and McGurk effect with lower computational cost. We think the artificial network motifs could be considered as some prior knowledge that would contribute to the multi-sensory integration of SNNs and provide more benefits for simulating the cognitive phenomenon. 

\end{abstract}


%
\IEEEpeerreviewmaketitle

\section{Introduction}

\IEEEPARstart{S}{piking} neural networks (SNNs) are considered as the third generation of artificial neural network (ANNs)~\cite{maass1997networks}, which are biologically plausible at network architectures, learning principles, and neuronal or synaptic types. These features in SNNs are far more complex and powerful than those used in ANNs~\cite{hassabis2017neuroscience}. Recently, it has been reported that even a single cortical neuron with dendritic branches needs at least a 5-to-8-layer deep neural network for finer simulations~\cite{RN763}, whereby non-differential spikes and multiply-disperse synapses make SNNs powerful on spatially-temporal information processing.

This paper highlights two important features of SNNs, which are also the most differences between SNNs and ANNs, specific network architecture and learning principles. The SNNs encode spatial information by fire rate and temporal information by spike timing, giving us hints and inspiration that SNNs are also powerful in integrating visual and auditory sensory signals.

For the architectures, specific cognitive topologies learned from evolution are highly sparse and efficient in SNNs \cite{luo2021architectures}, instead of pure densely-recurrent ones in counterpart ANNs. These fine micro-loops connections correspond to the cognitive functions. Many researchers also try their best to find out the biological nature of efficient multi-sensory integration by focusing more on the visual and auditory pathways in the biological brains \cite{Rideaux2021HowMN}.

For the learning principles, SNNs are more tuned by biologically-plausible plasticity principles, e.g., the spike timing-dependent plasticity (STDP)~\cite{zhang2017aplasticity}, short-term plasticity (STP)~\cite{zhang2018brain} (which further includes facilitation and depression), lateral inhibition, long-term potentiation (LTP) \cite{Teyler1987LongtermP}, long-term depression (LTD) \cite{Ito1989LongtermD}, Hebbian learning \cite{Kempter1999HebbianLA}, synaptic scaling, synaptic redistribution and reward-based plasticity~\cite{abraham1996metaplasticity}, instead of by the pure multi-step backpropagation (BP)~\cite{Rumelhart1986LearningRB} of errors in ANNs. The neurons in SNNs will not be activated until the membrane potentials reach thresholds, which makes them energy efficient. SNNs have been well applied on XOR problem~\cite{sporea2013supervised}, visual pattern recognition~\cite{diehl2015unsupervised,zeng2017improving,zhang2018plasticity,zhang2018brain,RN767}, auditory signal recognition~\cite{Jia2021NeuronalPlasticityAR}, probabilistic inference~\cite{soltani2010synaptic} and planning tasks~\cite{rueckert2016recurrent,ZhangDuzhen2021}. 

\begin{figure*}[htb]
\centering
\includegraphics[width=17cm]{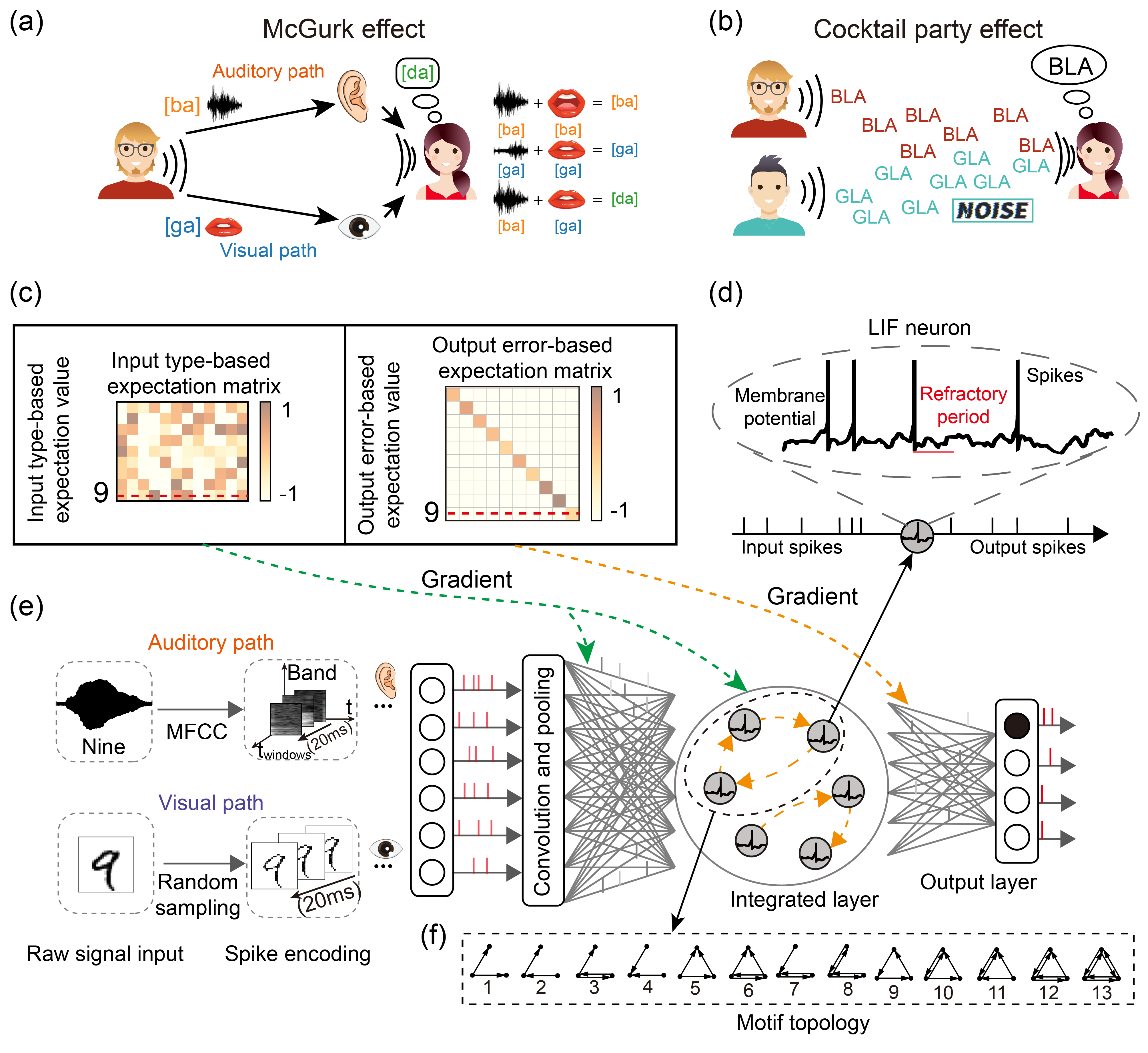}
\caption{The network structure for multi-sensory integration and two cognitive phenomena. (a), McGurk effect. New concepts arise when the receiver receives different audio-visual input information. (b), Cocktail party effect. When the receiver's brain focuses on one speaker, it filters out the sounds and noise from others. (c), Input and output transformation matrix of reward learning. (d), The spiking neuron with dynamic membrane potential. (e), The network of M-SNN on the single-sensory or multi-sensory integration task. (f), The example of artificial 3-node Motifs.}
\label{fig_MD_SNN}
\end{figure*}

For the two classic cognitive phenomena, the cocktail party effect describes the phenomenon that in a high-noise environment (e.g., environmental noise or other people's voices), when the listener focuses on one speaker, he would filter out the sounds from others as shown in Fig. \ref{fig_MD_SNN}(a). The McGurk effect introduced that the voice would be misclassified when the auditory stimulus and visual cues in the speakers are conflicting. A classic example of the McGurk effect describes the new concept [da] would be produced by the auditory input [ba] and visual cues [ga] as shown in Fig. \ref{fig_MD_SNN}(a).

In this paper, we focus more on the key features of SNNs at information integration, classification, and cognitive phenomenon simulated. The network motif (abbreviated as Motifs in this paper) \cite{Milo2002NetworkMS} in SNNs would be analyzed to reveal the essential functions of key meta circuits existing in SNNs and biological networks. Furthermore, we propose a method to mixture different motif structures and use them to simulate cognitive phenomenons, including cocktail party effects and McGurk effects. By comparing with networks without Motifs, network with Motif topologies can achieve higher classification accuracy in a simple cocktail party effect. In addition, the reward learning based on the clustering model can also be used to simulate the McGurk effect. Hence, a Motif-topology improved SNN (M-SNN) is proposed and then will be verified efficient (higher accuracy and better cognitive phenomenon simulation) on multi-sensory integration tasks. The main contributions of this paper can be summarized as follows:

\begin{itemize}
\item Specific spatial or temporal types of Motifs can improve accuracy at spatial or temporal classification tasks, respectively, making the multi-sensory integration easier by integrating two types of Motifs.
\item The proposed motif structure can better simulate cognitive phenomena and is validated in the cocktail party effect and McGurk effect.
\item In the process of network training for different simulation experiments, the M-SNN can achieve lower training computational cost than other SNN without Motif architectures which shown the M-SNN can achieve more human-like cognitive function with lower computational cost with the help of the prior knowledge of multi-sensory pathways and biological inspired reward learning method.
\end{itemize}

The remaining sections are organized as follows. Section \ref{sec:related} reviews the research about the learning paradigms and two classic cognitive phenomena. Section \ref{sec:methods} describes the pattern of Motifs, the SNN model with neuronal plasticity, and local and global learning principles. Section \ref{sec:experiment} verifies the convergence, advantage of M-SNN in simulation, and computational cost of the proposed M-SNN. Finally, a short conclusion is given in Section \ref{sec:conclusion}.

\section{Related works}\label{sec:related}

Many researchers are focusing on how to extract the motif topologies from a network. The feature of network could be reflected by the numbers of different motif topologies, which called the motif distribution. The first motif tool is mfinder, which implement the algorithm of full enumeration (randomly picking the edges from the graph and counting the probability of $n$-node subgraphs). Then the Mavisto, FANMOD, PGD, NetMODE have been published, which introduce the algorthms for analysing the graph more efficiently and are the more efficient tools for finding the Motifs. However, how to utilize the feature of Motifs is little researched. We have finished a shorter conference version of this paper appeared in an previous conference paper (ICASSP 2022)~\cite{RN801} that discussed the function of motif distributions in SNN of spatial and temporal tasks. In this paper, we further research the assistance of Motifs for better simulating the cognitive phenomenon including the McGurk effect and cocktail party effect.

For learning paradigms, besides biologically plausible principles (e.g., STDP, STP), some efficient algorithms have been proposed, such as ANN-to-SNN conversion (i.e., directly training ANNs with BP first and then equivalently converting to SNNs)~\cite{diehl2015fast}, proxy gradient learning (i.e., replacing the non-differential membrane potential at firing threshold by an infinite gradient value) \cite{lee2016training}, and temporal BP learning (e.g., SpikeProp) \cite{bohte2002error}. 

For solving the cocktail party effect, there are many effective models have been proposed. Wang proposed a ``Tune-In'' attention network with simulating the bottom-up and top-down system in human-beings, and utilise the self-training to train it for better performance in speech separation \cite{wang2021tune}. For the speaker extraction problem, Hao specially modeled the start and end information of the target speech to improve the recognition of the target speaker \cite{9413411}. In addition, the visual information are also introduced to enhance the recognition of the auditory signals \cite{Ephrat2018, chao2019speaker}. For McGurk effect, the researchers have found that the noise could influence the McGurk effect \cite{hirst2018threshold}, and the other pair of visual-auditory input was also found could influence the McGurk effect \cite{mcgurk1976hearing}. For the McGurk effect simulation, the unsupervised learning such as SOM have been introduced for modeling it\cite{gustafsson2014self}.

\section {Methods}\label{sec:methods}

\subsection{Spiking dynamics}

The leaky integrated-and-fire (LIF) neuron model is biologically plausible and one of the simplest models to simulate spiking dynamics. It includes non-differential membrane potential and refractory period, as shown in Fig. \ref{fig_MD_SNN}(d). The LIF neuron model simulates the neuronal dynamics by the following steps. 

First, the dendritic synapses of the postsynaptic LIF neuron will receive presynaptic spikes and transfer them to postsynaptic current ($I_{syn}$). Second, the postsynaptic membrane potential will be leaky or integrated affected by its experience. The classic LIF neuron model is shown as the following Equation (\ref{equa_lif}).

\begin{equation}
\tau_m\frac{dV(t)}{dt}=-\left(V(t)-V_L \right )-\frac{g_{E}}{g_L}\left(V(t)-V_{E} \right )+ \frac{I_{syn}}{g_L}
\textbf{,}
\label{equa_lif}
\end{equation}

where $V(t)$ represents the dynamical variable of membrane potential with time $t$, $dt$ is the minimal simulation time slot (set as 0.01ms), $\tau_m$ is the integrative time period, $g_L$ is the leaky conductance, $g_E$ is the excitatory conductance, $V_L$ is the leaky potential, $V_E$ is the reversal potential for excitatory neuron, and $I_{syn}$ is the input current received from the synapses in the previous layer. We set values of conductance ($g_E,g_L$) to be $1$ in our following experiments for simplicity, as shown in Equation (\ref{equa_recurrent_SNN}). 

Third, the postsynaptic neuron will generate a spike once its membrane potential $V(t)$ reaches the firing threshold $V_{th}$. At the same time, the membrane potential $V$ will be reset as the reset potential $V_{reset}$, shown as the following Equation (\ref{equalifall}).

\begin{equation}
\text { if }\left(V(t)>V_{t h}\right)\left\{
\begin{array}{l}
V(t)=V_{reset} \\
T_{ref}=T_{0}
\end{array}\right.
\textbf{,}
\label{equalifall}
\end{equation}

where the refractory time $T_{ref}$ will be extended to a larger predefined $T_0$ after firing.

In our experiments, the three steps for simulating the LIF neurons were integrated into the Equation (\ref{equa_lif}).

\begin{equation}
    \begin{aligned}
  C \frac{dV_i(t)}{dt}=g\left(V_i(t)-V_{rest}\right)\left(1-S_i(t)\right)+\sum_{j=1}^NW_{i,j}X_j(t)
    \end{aligned}
    \label{equa_lif} \text{,}
\end{equation}

where $C$ is the capacitance parameter, $S_i(t)$ is the firing flag of neuron $i$ at timing $t$, $V_i(t)$ is the membrane potential of neuron $i$ at timing $t$, $V_{rest}$ is the resting potential, $W_{i,j}$ represent the synaptic weight between the neuron $i$ and $j$.

\subsection{The Motif topology and distribution}

In the past research, the $n$-node ($n\geq 2$) meta Motifs have been proposed. Here, we use the typical 3-node Motifs to analyze the networks, which has been widely used in biological and other systems~\cite{RN800,Milo2002NetworkMS,2012Information}. All the 13 types of 3-node Motifs were showen in Fig. \ref{fig_MD_SNN}(f). In SNNs, the Motifs were represented by the motif masks that were applied into the recurrent connection at the hidden layer.

The typical motif mask is a matrix padded with 1 or 0, where 1 and 0 represent the connected and non-connected pathways, respectively. We introduce the motif circuits into the hidden layer, and the motif mask in the $r$-dimension hidden layer $l$ at time $t$ is represented as the $M_t^{r,l}$ as shown in Equation (\ref{hard_motif}).

\begin{equation}
\begin{aligned}
    M_t^{r,l} = \begin{bmatrix}
 f(m_{1,2}) & \cdots & f(m_{1,r}) \\
 \vdots  &  \ddots & \vdots \\
 f(m_{r,1}) & \cdots & f(m_{r,r})
\end{bmatrix}\\
\end{aligned}
    \label{hard_motif} \text{,}
\end{equation}

where $f(\cdot)$ is the indicator function. Once the variable in $f(\cdot)$ satisfied the conditions, the function value would be set as one, otherwise zero. $m_{i,j}, (i,j=1,\cdots r)$ are elements of synaptic weight $W_t^{r,l}$.

The network motif distribution is defined as the function with the independent variable of the network motif index and the dependent variable of frequency. The function could be obtained by counting the numbers of different network motif types. The network motif distribution can be obtained by enumerating the 3-node connected subgraphs of the original network and counting the number of them. In this paper, we use the FANMOD to calculate the numbers of motifs.

\subsection{Lateral connections of spiking neurons}

The lateral and sparse connections between LIF neurons are usually designed to generate network-scale dynamics. As shown in Fig. \ref{fig_MD_SNN}(e), we designed a four-layer SNN architecture containing input layer (for pre-encoding visual and auditory signals to spike trains), convolutional layer, multi-sensory integration layer, and readout layer. The weights of the synapses among neurons in the same hidden layer are learnable, but the connection state of the synapses are controlled by the Motif masks. The membrane potentials in the hidden multi-sensory-integration layer are updated by both feed-forward potential and recurrent potential, shown as the following Equation (\ref{equa_recurrent_SNN}):

\begin{equation}
    \left\{\begin{array}{l}
    \begin{matrix}
        S_i(t) = S_i^f(t) + S_i^r(t)
    \end{matrix}\\
    \begin{matrix}
        V_i(t) = V_i^f(t) + V_i^r(t)
    \end{matrix}\\
    \begin{matrix}
        C\frac{dV_i^f(t)}{dt}=&g(V_i(t)-V_{rest})(1-S(t))\\
        &+\sum_{j=1}^NW^f_{i,j}X_j(t)
    \end{matrix}\\
    \begin{matrix}
        C\frac{dV_i^r(t)}{dt}=\sum_{j=1}^NW^r_{i,j}S(t)\cdot M_{t}^{r,l}
    \end{matrix}\\
    \end{array}\right.
    \textbf{,}
    \label{equa_recurrent_SNN}
\end{equation}

where $C$ is the capacitance parameter, $S_i(t)$ is the firing flag of neuron $i$ at timing $t$, $V_i(t)$ is the membrane potential of neuron $i$ at timing $t$ that incorporates feed-forward $V_i^f(t)$ and recurrent $V_i^r(t)$, $V_{rest}$ is the resting potential, $W_{i,j}^f$ is the feed-forward synaptic weight from the neuron $i$ to the neuron $j$, $W_{i,j}^r$ is the recurrent synaptic weight from the neuron $i$ to the neuron $j$. $M_{t}^{r,l}$ is the mask that incorporates Motif topology to influence the feed-forward propagation further. The historical information is stored in the forms of recurrent membrane potential $V_i^r(t)$, where spikes are generated after potential reaching a firing threshold, shown as the follow Equation (\ref{equa_SnnUnit1}).

\begin{equation}
    \left\{\begin{array}{l}
    V_{i}^{f}(t)=V_{reset}, S^{f}=1 \quad if (V_{i}^{f}(t)=V_{th}) \\
    V_{i}^{r}(t)=V_{reset}, S^{r}=1 \quad if \left(V_{i}^{r}(t)=V_{t h}\right) \\
    S^{f}(t)=1 \quad i f\left(t-t_{s^{f}}<\tau_{r e f}, t \in\left(1, T_{1}\right)\right) \\
    S^{r}(t)=1 \quad i f\left(t-t_{s^{r}}<\tau_{r e f}, t \in\left(1, T_{2}\right)\right)
    \end{array}\right.
    \textbf{,}
    \label{equa_SnnUnit1}
\end{equation}

where $V_i^f(t)$ is the feed-forward membrane potential, $V_i^r(t)$ is the recurrent membrane potential, $S^f(t)$ and $S^r(t)$ are spike flags of feed-forward and recurrent membrane potentials, respectively, $V_{reset}$ is reset membrane potential.

\subsection{The local neuronal plasticity and learning principle}

\subsubsection{Neuronal plasticity with adaptive threshold}

Neuronal plasticity puts more emphasis on spatially-temporal information processing by considering the inner neuron dynamic characteristics \cite{Jia2021NeuronalPlasticityAR}, which differs from the traditional synaptic plasticity such as STP and STDP. The neuronal plasticity for SNNs approaches the biological network and improves the learning power of the network. 
Rather than being a constant value, the firing threshold is set by an ordinary differential equation shown as follows:

\begin{equation}
    \frac{da_i(t)}{dt} = (\alpha-1) a_i(t) + \beta (S^f(t)+S^r(t))
    \label{equa_adaptive} \text{,}
\end{equation}

where $S^f(t)$ is the input spikes from the feed-forward channel. $S^r(t)$ is the input spikes from the recurrent channel. $a_i(t)$ is the dynamic threshold, which has an equilibrium point of zero without input spikes or $-\frac{\beta}{\alpha-1}$ with input spikes $S^f+S^r$ from the feed-forward and recurrent channels. Therefore, the membrane potential of adaptive LIF neurons is updated as follows:

\begin{equation}
    \begin{aligned}
  C \frac{dV_i(t)}{dt}=g\left(V_i(t)-V_{rest}\right)\left(1-S^f(t)-S^r(t)\right)\\
  +\sum_{j=1}^NW_{i,j}X_j(t)- \gamma a_i(t)
    \end{aligned}
    \label{equa_adaptive2} \text{,}
\end{equation}

where the dynamic threshold $a_i(t)$ is accumulated during the period from the resetting to the membrane potential firing and finally attain a relatively stable value $a_i^*(t)=\frac{\beta}{1-\alpha}(S^f(t)+S^r(t))$. Because of the $- \gamma a_i(t)$, the maximum firing threshold could reach up to $V_{th}+\gamma a_i(t)$.

We make $\alpha=0.9$, $\beta=0.1$, $\gamma=1$. Accordingly, the stable $a^*(t)=0$ for no input spikes, $a^*(t)=1$ for one input spike, $a^*(t)=2$ for input spikes from two channels. When $a_i(t)<(S^f(t)+S^r(t))$, the threshold $a_i(t)$ will increase, otherwise, the threshold $a_i(t)$ will decrease. It is clear that the threshold will change in the process of the neuron's firing, and as the firing frequency of the neuron increases, the threshold will also elevate, or vice versa.

\subsubsection{Local synaptic plasticity with gradient approximation}

The membrane potential at the firing time is a non-differential spike, so local gradient approximation (pseudo-BP)~\cite{Tuningzhang2020} is usually used to make the membrane potential differentiable by replacing the non-differential part with a predefined number, shown as follows:

\begin{equation}
    Grad_{local}=\frac{\partial S_i(t)}{\partial V_i(t)}=\left\{\begin{array}{cc}
1 & i f\left(\left|V_i(t)-V_{t h}\right|<V_{win}\right) \\
0 & else
\end{array}\right.
\textbf{,}
\label{equa_local}
\end{equation}

where $Grad_{local}$ is the local gradient of membrane potential at the hidden layer, $S_i(t)$ is the spike flag of neuron $i$ at timing $t$, $V_i(t)$ is the membrane potential of neuron $i$ at timing $t$, $V_{th}$ is the firing threshold. This approximation makes the membrane potential $V_i(t)$ differentiable at the spiking time between an upper bound of $V_{th}+V_{win}$ and a lower bound of $V_{th}-V_{win}$.

\subsection{The global principle of reward learning}

The reward propagation has been proposed in our previous work \cite{Tuningzhang2020}. As shown in Fig. \ref{fig_MD_SNN}(c), the gradient of the hidden layer in training are generated from the input type-based expectation value and output error-based expectation value by transformed matrix (input type-based expectation matrix and output error-based expectation matrix), respectively, then the gradient signal will is directly given to all hidden neurons without layer-to-layer backpropagation, shown as follows: 

\begin{equation}
\left\{\begin{array}{l}
Grad_{R_l}=B_{rand}^{f,l}\cdot R_{t}-h^{f,l} \\
Grad_{R_L}=B^{f,L}\cdot e^{f,L} \\
\Delta W_{t}^{f,l}=-\eta^{f}(Grad_{R_l}) \\
\Delta W_{t}^{r,l}=-\eta^{r}\left(Grad_{t+1}+Grad_{R_l}\right)\cdot M_{t}^{r,l} \\
\Delta W_{t}^{f,L}=-\eta^{f}(Grad_{R_L})
\end{array}\right.\text{,}
\label{equa_r}
\end{equation}

where $h^{f,l}$ is the current state of layer $l$, $R_t$ is the predefined input-type based expectation value. A predefined random matrix $B_{rand}^{f,l}$ is designed to generate the reward gradient $Grad_{R_l}$. $Grad_{R_L}$ is gradient value of the last hidden layer, $B^{f,L}$ is the predefined identity matrix, and $e^{f,L}$ is the output error. $W_t^{f,l}$ is the synaptic weight at layer $l$ in feed-forward phase, $\Delta W_t^{r,l}$ is the recurrent-type synaptic modification at layer $l$ which is defined by both $Grad_{R_l}$ by reward learning and $Grad_{t+1}$ by iterative membrane-potential learning, and the $Grad_{t+1}$ means the gradient obtained at $t+1$ moment \cite{Werbos1990BackpropagationTT}. The $M_{t}^{r,l}$ is the mask that incorporates Motif topology to further influence the propagated gradients.

\subsection{Network learning and Motif-mask integration}

\begin{figure}[htb]
\centering
\includegraphics[width=8.5cm]{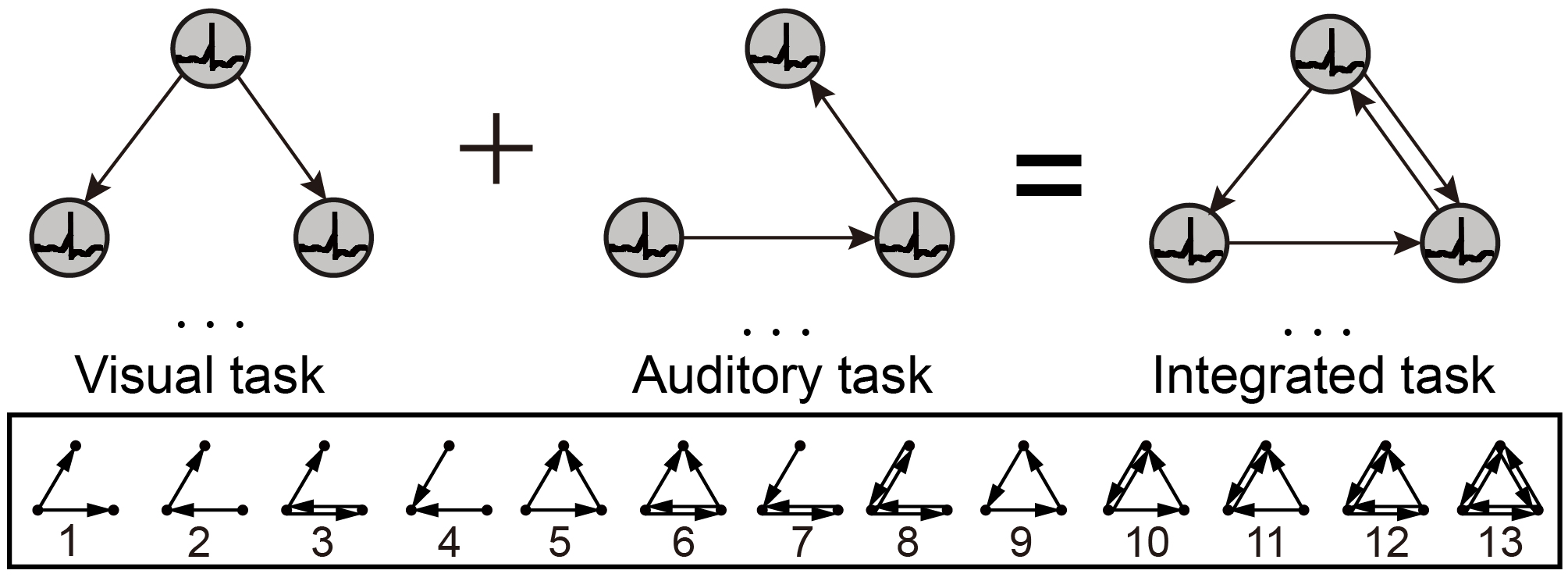}
\caption{Schematic diagram of integrating Motif masks.}
\label{fig_ProcessOfIntegration}
\end{figure}

We propose a multi-sensory integration algorithm for integrating Motif masks with different types learned from visual or auditory classification tasks. The visual and auditory Motif masks are trained by the BP algorithm from the MNIST and TIDigits dataset. Then the spatial and temporal Motif masks would be integrated by our proposed method. Fig. \ref{fig_ProcessOfIntegration} showed one example of generate the integrated Motif, the integrated Motif is the union of the visual and auditory connections.

\subsection{The learning procedure of M-SNN}

The overall learning procedures of the M-SNN were shown in Algorithm \ref{algorithm1}, including raw signal encoding, local-and-global synaptic weight learning and motif structure integration.

\begin{algorithm}
\caption{The M-SNN algorithm.}
\label{algorithm1}
\begin{algorithmic}
\State 1. Initialize the network by resetting weights and all related parameters. e.g., initial membrane potential $V_i$, simulation time $T$, learning rates $\eta=\eta^f=\eta^r$.
\State 2. Encode raw numbers of datasets to spike trains.
\State 3. Learn the synaptic weights $w_{ij}$ and Motif masks $M_t^{r,l}$ by BP~\cite{Rumelhart1986LearningRB} in two single-sensory tasks to get the spatial mask $M_t^{r,l}\left(s\right)$ and temporal mask $M_t^{r,l}\left(t\right)$.
\State 4. Synthesize Motif masks and train the synaptic weight $w_{ij}$ on multi-sensory integration tasks.
\State 4.1 Synthesize the integrated masks $M^{r,l}_{t}$ from spatial and temporal masks, where $M^{r,l}_{t}= M^{r,l}_{t}(s) \cup M^{r,l}_{t}(t)$.
\State 4.2 Initialize a new network and add the Motif mask $M_t^{r,l}$. 
\State 4.3 Only learn the synaptic weight $w_{ij}$ with local Pseudo-BP and global reward learning~\cite{Tuningzhang2020}.
\State 5. Test the performance of SNNs using these new masks in the multi-sensory classification tasks and simulate the cocktail party effect and McGurk effect.
\end{algorithmic}
\end{algorithm}

\section{Experiments}\label{sec:experiment}

\subsection{Visual and auditory Datasets}

The MNIST dataset~\cite{lecun1998mnist} was selected as the visual sensory dataset, containing 70,000 28$\times$28 one-channel gray images of handwritten digits from zero to nine. Among them, 60,000 images are selected for training, while the remaining 10,000 ones are left for testing. The TIDigits dataset \cite{RN798} was selected as the auditory sensory dataset, containing 4,144 spoken digit recordings from zero to nine, corresponding to those in the MNIST dataset. Each recording was sampled as 20KHz for around one second. Some examples were shown in Fig. \ref{fig_MD_SNN}(e).

\subsection{Experimental configurations}
\label{subsection:experimental}

We built the SNN in Pytorch and trained on TITAN Xp GPU. The network architectures for MNIST and TIDigits were the same, containing one input encoding layer, one convolutional layer (with a kernel size of 5$\times$5), one full-connection or integrated layer (with 200 LIF neurons), and one output layer (with ten output neurons). The capacitance $C$ was 1$\mu \text{F/cm}^2$, conductivity $g$ was 0.2 nS, time constant $\tau_{ref}$ was 1 ms, resting potential $V_{rest}$ was equal to reset potential $V_{reset}$ with 0 mV. The learning rate was $1e$-$4$, the firing threshold $V_{th}$ was 0.5 mV, the simulation time $T$ was set as 28 ms, the gradient approximation range $V_{win}$ aws 0.5 mV.

\begin{figure*}[htb]
\centering
\includegraphics[width=17cm]{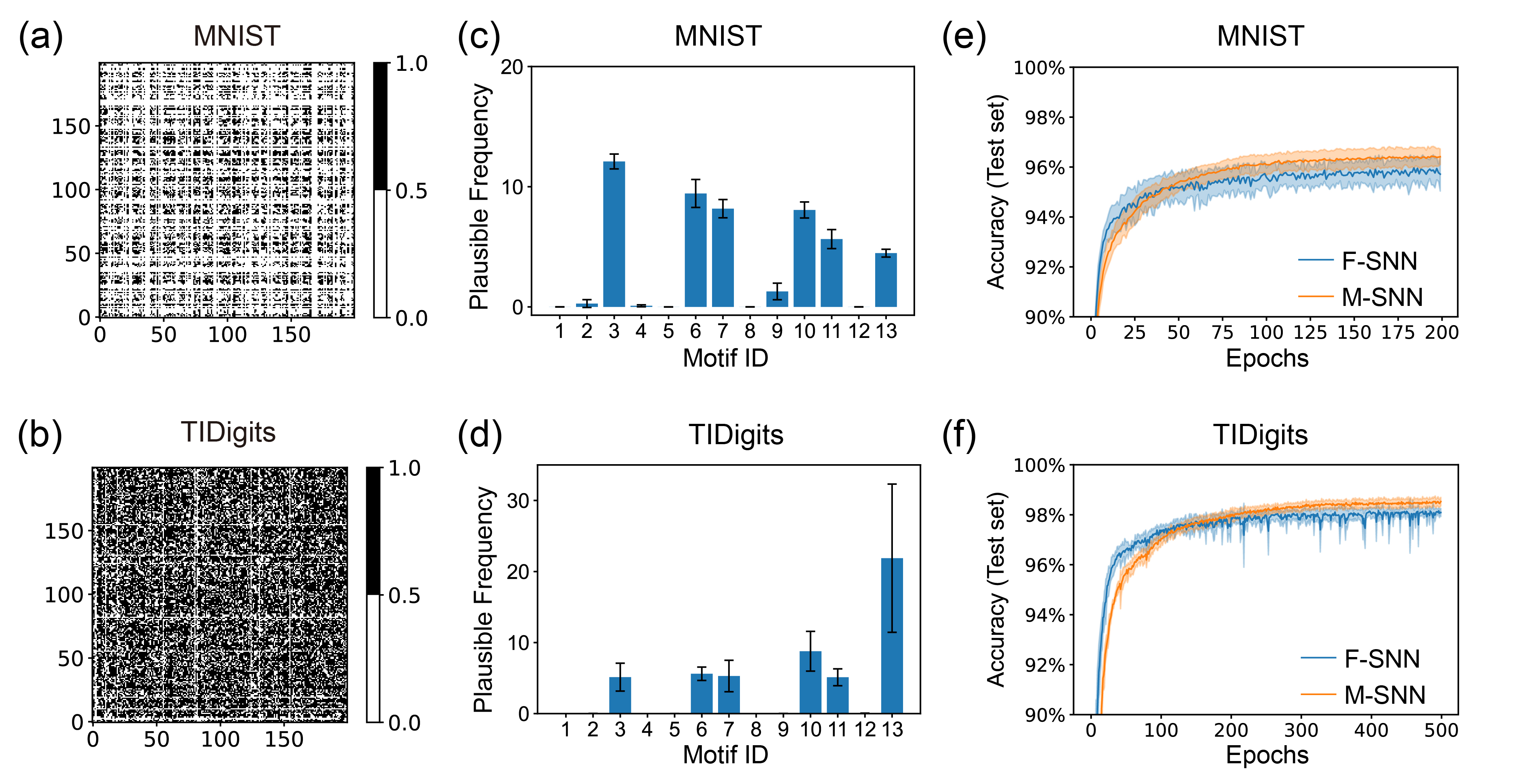}
\caption{Network convergence with SNNs using Motifs in different datasets. (a, b), Motif masks of MNIST (a) and TIDigits (b) after training. (c, d), Plausible Frequency of Motif distributions of MNIST (c) and TIDigits (d) datasets after training. (e, f), Convergence curve of classification task of MNIST (e) and TIDigits (f) datasets.}
\label{fig_motif}
\end{figure*}

As shown in Fig. \ref{fig_MD_SNN}(e), for the visual dataset, before being given to the input layer, the raw data were encoded to spike trains first by comparing each number with a random number generated from Bernoulli sampling at each time slot of the time window $T$. For the auditory dataset, the input data would be transformed to the frequency spectrum in the frequency domain by MFCC (Mel frequency cepstrum coefficient) first, and then the spectrum would be split according to the time windows; finally, the sub-spectrum would be randomly sampled and encoded to spike trains. 

There are two SNNs are concluded in our experiment as follows:

\begin{itemize}
    \item M-SNN. The motif mask is generated randomly and then updated during the learning of synaptic weights in a feedforward SNN (F-SNN).
    \item F-SNN. The standard feed-forward SNN without motif masks plays as the control algorithm for comparing M-SNN.
\end{itemize}

\subsection{Analysis of spatial and temporal Motif topology during learning}

The visual and auditory Motif masks were shown in Fig \ref{fig_motif}, which were trained from the MNIST and TIDigits datasets, respectively. The generated visual Motif mask after training was shown in Fig. \ref{fig_motif}(a,b), in which the black dot in the visualization of the Motif mask meant there was a connection between the two neurons shown at the X-axis and Y-axis, and the white dot meant not. This result showed that the visual Motif mask connections were sparse, where only about half neurons were connected. For the temporal TIDigits dataset, the generated temporal Motif mask after training was shown in Fig. \ref{fig_motif}(b), where the learned Motif mask was denser than that on visual MNIST in \ref{fig_motif}(a). It is important that the denser temporal Motifs correspond to the biological findings \cite{vinje2000sparse,hromadka2008sparse}. These differences between spatial and temporal Motif masks indicated that the network needed a more complex connection structure to deal with the sequential information. 

For further analysis the Motif structures, we used the ``Plausible Frequency'' instead of the standard frequency to calculate the significant Motifs after comparing them to the random networks. The ``Plausible Frequency'' was defined by multiplying the occurrence frequency and $1-P$, where the $P$ was the P-value of a selected Motif after comparing it to 2,000 repeating control tasks with random connections. The ``repeating control tasks'' meant generating many matrixes (e.g., 2000) that each element was sampled from a uniformly random distribution. And $P$-value was an index that showed the statistical significance of the concerning results, and a lower $P$-value indicated more plausible. The Motif distributions corresponding to the Motif masks were shown in Fig. \ref{fig_motif}(c,d), where the spatial or temporal Motifs were very differently distributed. For spatial Motifs, the 3rd, 6th, 7th, 10th, and 11th Motifs were all prominent, while for temporal Motifs, the 13th Motif was the most prominent. The plausible frequency shown the key meta neuronal circuit in the Motif mask that excluded the interference from random connection. Fig. \ref{fig_motif}(e,f) shown that M-SNN networks with Motif topologies could maintain convergence, and after some training epochs, the accuracy of M-SNN was significantly higher than the accuracy of F-SNN.

\begin{figure*}[htb]
\centering
\includegraphics[width=17cm]{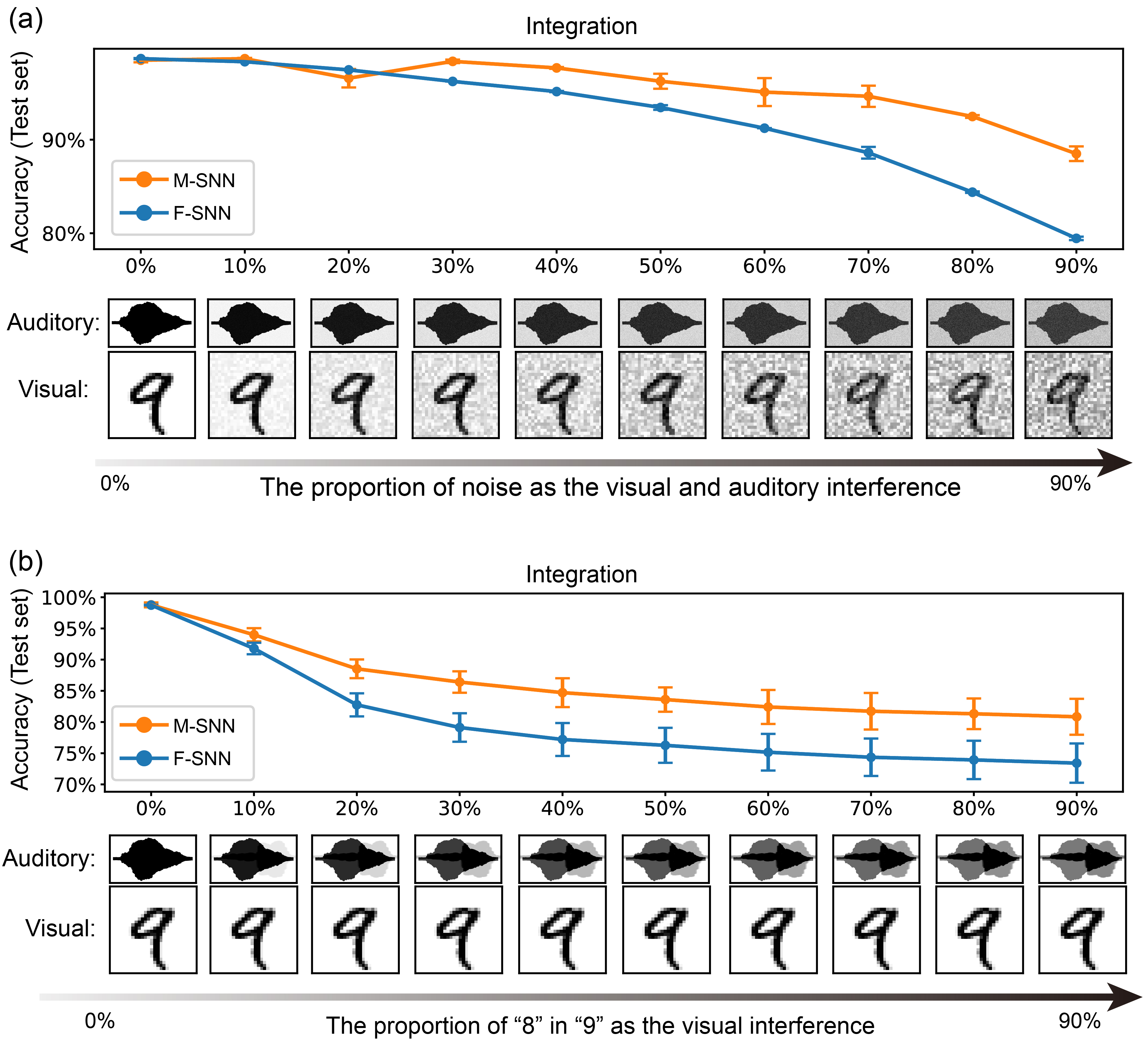}
\caption{Simulation of cocktail party effect. (a), A simulation and results in which both visual and auditory inputs are interfered. (b), A simulation and results in which only the voice are interfered. All figures are averaged over five repeating experiments with different random seeds.}
\label{fig_curve}
\end{figure*}

\begin{figure*}[htb]
\centering 
\includegraphics[width=17cm]{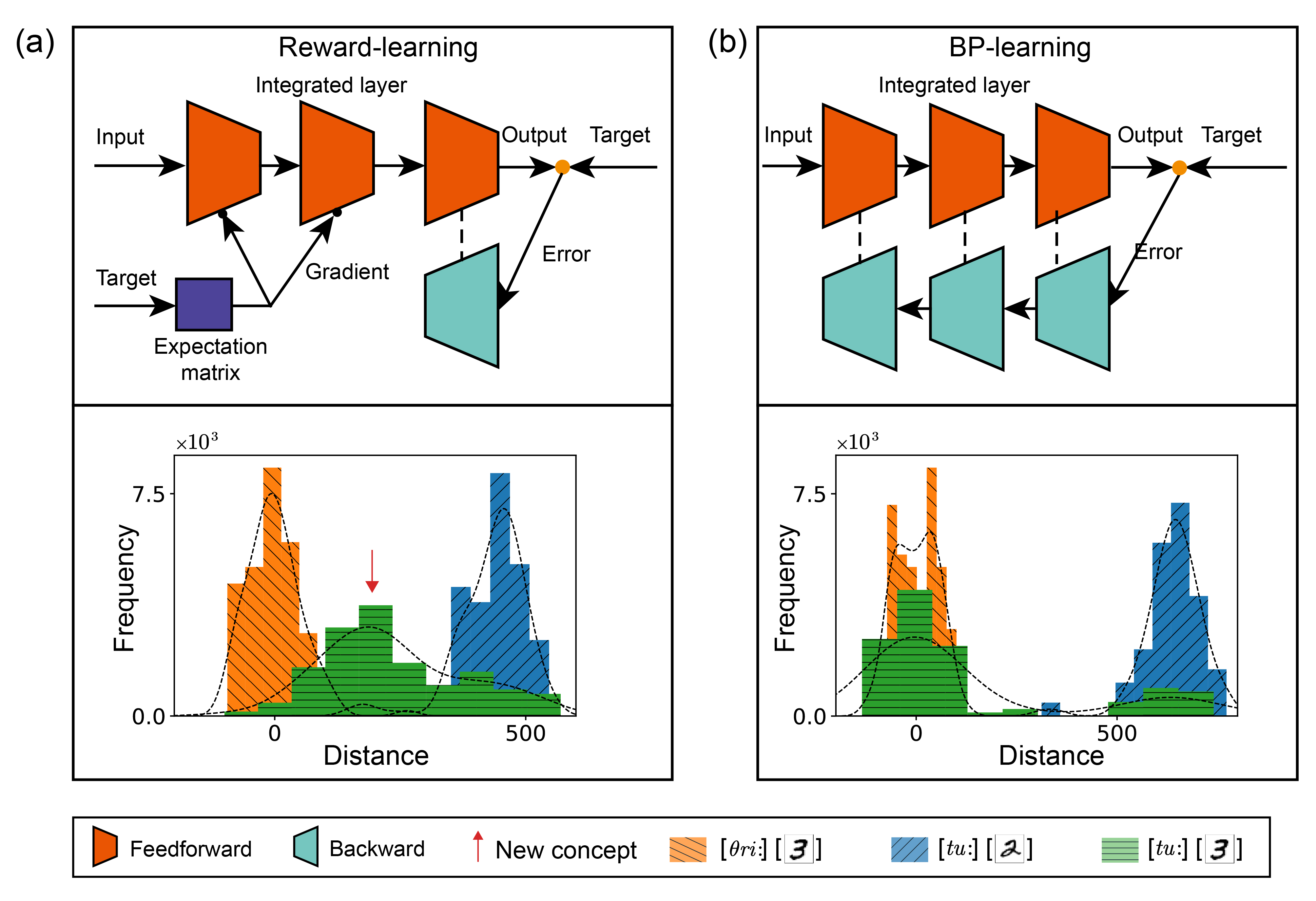}
\caption{Simulation of McGurk effect. (a), Reward learning and simulation of McGurk effect. Top panel: Schematic of reward learning. Bottom panel: Distribution in integrated layer after reward learning of different combinations of input. (b), BP learning and simulation of McGurk effect. Top panel: Schematic of reward learning. Bottom panel: Distribution in integrated layer after BP learning of different combinations of input.}
\label{fig_motif}
\end{figure*}

\begin{figure*}[htb]
\centering
\includegraphics[width=17cm]{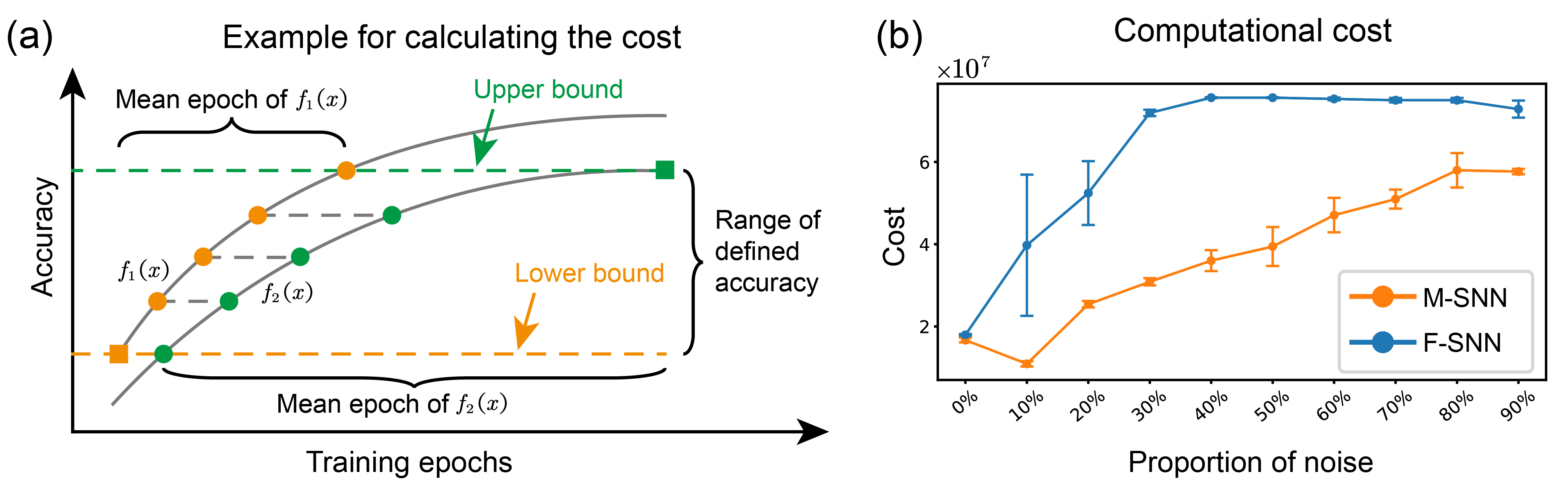}
\caption{M-SNN in training for lower computational cost. (a), Schematic diagram depicting how to calculate the mean epoch during training. (b), The computational cost of network training under different proportion of noise.}
\label{fig_cost2}
\end{figure*}

\subsection{M-SNN contribute to solve the cocktail party effect}

We modeled two scenes to simulate the simplified cocktail party effect. In modeling with the two-channel network, we used the MNIST dataset representing the visual input and the TIDigits dataset for phonetic input. The first scene simulated a cocktail party effect where visual and auditory inputs were disrupted in a noisy environment. The second scene simulated a cocktail party effect where only the human voice was disturbed.

\subsubsection{Both visual and auditory inputs are interfered}
Both visual and auditory inputs were interfered with in a simulated environment: There were a large number of people, and talking sounds from other people form a relatively large noise, and the environment was dark. The listener observed that the speaker's facial expression and lip movement information significantly interfered.

When training the network, we added the same proportion of noise to the training and test datasets. In the process of simulation, we used the method of superimposing random numbers between [0,1] into the image or speech input to simulate the interference effect of noise. With the different values of the added random numbers, different interference effects were formed, ranging from 0\% to 90\%, and the influence gradually increased. As shown in Fig. \ref{fig_curve}(a), when the influence of noise was relatively low, whether to add Motifs into the network had little effect on the experimental results (98.50 $\pm$ 0.22\% for the network with Motifs, 98.67 $\pm$ 0.01\% for F-SNN). With the increasing in noise ratio, the recognition ability of the network to the input target signal decreased gradually. When the proportion of noise was added into 90\%, the accuracy of the M-SNN was 88.50 $\pm$ 0.78\% which was markedly higher than the accuracy of F-SNN (79.45 $\pm$ 0.2\%). The higher accuracy indicated that the Motifs in M-SNN had a better effect on solving the cocktail party effect. This better effect would become more evident with the increasing in the noise ratio. In these situations, the maximal increased accuracy was 9.1\% when the proportion of noise was 90\%.

\subsubsection{Only the voice are interfered}

Only the voice interfered with the situation meant that only auditory was disturbed by other sounds, while visual information that the listener received from other speakers was not disturbed. 

When training the network, we used the MNIST and TIDigits datasets without noise. In the simulation process, we used ``8'' from the natural human voice instead of a random number as noise interference and kept the same image input. In the case of a few other interfering sounds, the effect of M-SNN on maintaining accuracy was not significant. However, with the increase in the proportion of different interfering sounds, the impact of M-SNN on maintaining the recognition of the network was becoming more and more significant. When the noise ratio reached 90\%, the recognition accuracy of M-SNN got 80.84 $\pm$ 2.88 \%, while the F-SNN could only reach the accuracy of 73.41 $\pm$ 3.15\%. In these situations, the maximal increased accuracy was 7.5\% when the proportion of ``8'' was 40\%.

\subsection{M-SNN for explainable McGurk effect}

The McGurk effect described the psychological phenomenon that when human speech input and image input were inconsistent, most people would judge the input as neither a speech label nor a visual label. It had been shown that for adults, the error rate in judging inconsistent audio-visual input as novel concepts was more than 90\%. For example, when the speech input was [ba] and the visual input was [ga], a new concept [da] was generated. When the speech input was [ga] and the visual input was [ba], a new concept [da] was generated.

\subsubsection{Simulated the McGurk effect}

As shown in Fig. \ref{fig_motif}, we trained the model using reward learning and BP learning, respectively. The reward learning used the target to generate the gradient without the layer-to-layer gradient backpropagation algorithm of BP learning. After training, the inconsistent audio-visual information would be fed into the network. In the integrated layer, we used TSNE to reduce the dimensionality of the high-dimensional features to obtain the representation of high-dimensional features in low-dimensional space. During the simulation, we used handwritten digit images [2],[3] to represent the visual input, while speech digits [$tu$:],[$\theta ri$:] were used to represent the corresponding pronunciation. 

The histogram in the lower part of Fig. \ref{fig_motif} shown the distribution of samples with different labels in the integration layer. For the learning results of BP learning, there were two prominent feature distributions: [$\theta ri$:,3] and [$tu$:,2]. However, for the learning results of M-SNNs, a clear feature distribution of [$tu$:,3] emerged between the distributions of [$\theta ri$:,3] and [$tu$:,2]. This distribution corresponding to [$tu$:,3] characterized the new concept.

\subsubsection{Feasibility analysis}

Instead of using the error back-propagation algorithm to learn the network weights, the reward learning algorithm used the expectation matrix to generate gradients for modifying the network weights, which was closer to a clustering model. The target corresponding to different inputs would be encoded by the expectation matrix and then directly generated the gradient to guide the integrated layer weight modification. The modified network would guide the feature of inputs to different feature spaces in the integration layer. When the input audio-visual was inconsistent, there would be a new region in the cluster space of the M-SNN, whereby the simulation of the McGurk effect could be realized.

\subsection{Lower computational cost for M-SNN during training}

We calculated the computational cost of training for different proportions of noise, as shown in Fig. \ref{fig_cost2}(b). The different kinds of noise represented different experiment situations. The 0\% noise in the training set was used for simulating the McGurk effect, and the other proportion s of noise was used for simulating the cocktail party effect in previous Sections.

We referred to the method in paper \cite{RN767} to calculate the computational cost of training of the network for algorithm $i, (i$=$1,2)$, where the average training cost of the network was represented by the average epoch multiplied by the number of parameters of the network. Schematic for the mean epoch was shown in Fig. \ref{fig_cost2}(a), and the equation was shown as follows:

\begin{equation}
Cost_i=\frac{1}{N} \sum_{l=1}^{N} \text{Argmin}_i \left(f_i(x)=Acc_{l}\right) \times O(n)_i \text{,}
\label{equa_cost}
\end{equation}

where $\text{Argmin}_i(\cdot)$ is the argument when $\cdot$ is the minimum, $f_i(x)$ is accuracy function of training epoch $x$, $Acc_l$ is the selected upper bound of accuracy in $[f_1(x), f_2(x)]$, $O(n)_i$ is algorithmic complexity of algorithm $i$.

The results of M-SNN and F-SNN computational cost were shown in Fig. \ref{fig_cost2}. With the increase of the proportion of noise in the dataset, the computational cost in the training of M-SNN gradually increased, while the training cost of F-SNN would increase to a maximum value and tend to be stable. Therefore, the increased noise ratio brought higher computational to the network. In addition, Motifs in M-SNN could save network training computational cost (the training cost convergence curves of M-SNN was always below the convergence curves of F-SNN), and M-SNN achieved the maximum cost-saving ratio of 72.6\% when the noise ratio was 10\%. When the noise ratio reached 30\%, M-SNN achieved the most significant absolute cost savings of $4.1 \times 10^7$.

\section{Conclusion}\label{sec:conclusion}

In this paper, we proposed a model of Motif-topology improved SNN (M-SNN), exhibiting three main important features. First, M-SNN could simulate the cocktail party effect with a better effect. Compared with the common F-SNN, M-SNN had a better function of filtering noise and other speaker sounds in different proportions. Second, compared with SNN with BP learning, M-SNN with reward learning was helpful in simulating the McGurk effect, and M-SNN with auditory and visual Motifs could better explain the McGurk effect. Third, compared with F-SNN, M-SNN had lower computational cost of training in simulating the environment with different noise ratios, and the maximum computational cost saving ratio is 72.6\%. 

The deeper analysis of the Motifs helped us understanding more about the key functions of the structures in SNNs. This inspiration from Motifs described the sparse connection in the cell assembly that revealed the micro-scale importance of the structures. Motif topologies were patterns for describing the topologies of a system (e.g., biological cognitive pathways), including the $n$-node meta graphs that uncover the bottom features of the networks. We found the biological Motifs were beneficial for improving the accuracy of networks in visual and auditory data classification. Significantly, the 3-node Motifs were typical and concise, which could assist in analyzing the function of different network modules.

We think the research on the variability of Motifs will give us more hints and inspirations for a better network, and the simulation of different cognitive functions by SNNs with biologically plausible Motifs has much in store in the future.

\section*{Acknowledgment}
The authors would like to thank Hongxing Liu for his previous assistance with the discussion. This study is supported by the National Key R\&D Program of China (Grant No. 2020AAA0104305), the Strategic Priority Research Program of the Chinese Academy of Sciences (Grant No. XDB32070100, XDA27010404), and the Shanghai Municipal Science and Technology Major Project. The source code of the models and experiments can be found at https://github.com/thomasaimondy/Motif-SNN. The authors declare that they have no competing interests.

\ifCLASSOPTIONcaptionsoff
\newpage
\fi

\bibliographystyle{IEEEtran}
\bibliography{thomas}

\newpage
\vspace{-80 mm}
\begin{IEEEbiography}[{\includegraphics[width=1in,height=1.0in,clip,keepaspectratio]{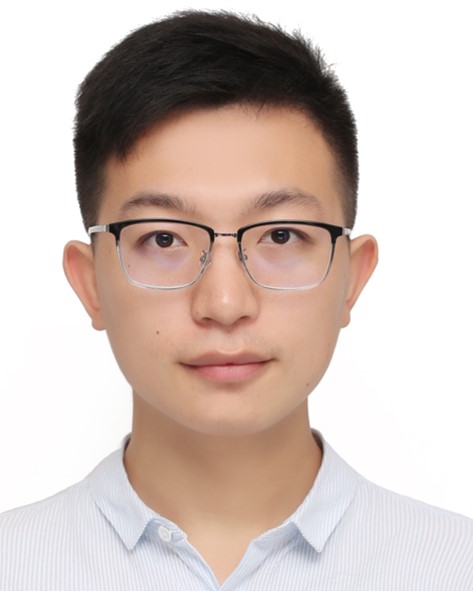}}]{Shuncheng Jia}
is a Ph.D. candidate in both the Institute of Automation Chinese Academy of Sciences and the University of Chinese Academy of Sciences. His current interests include theoretical research on neural dynamics, auditory signal processing, and Spiking Neural Networks.
\end{IEEEbiography}

\vspace{-80 mm}
\begin{IEEEbiography}[{\includegraphics[width=1in,height=1.0in,clip,keepaspectratio]{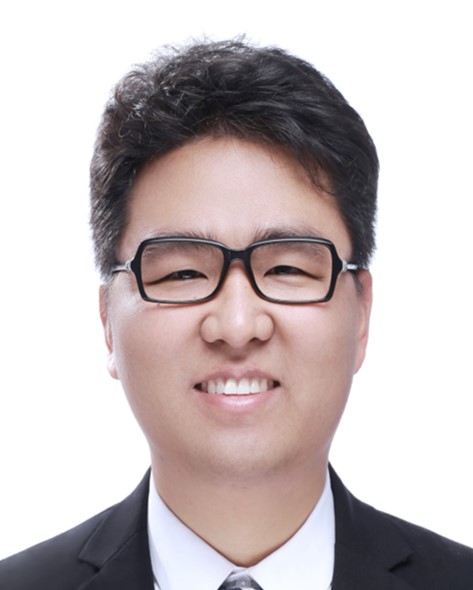}}]{Tielin Zhang}
received the Ph.D. degree from the Institute of Automation Chinese Academy of Sciences, Beijing, China, in 2016. He is an Associate Professor in the Research Center for Brain-Inspired Intelligence, Institute of Automation, Chinese Academy of Sciences, Beijing, China. His current interests include theoretical research on neural dynamics and Spiking Neural Networks (more information is in https://bii.ia.ac.cn/$\sim$tielin.zhang/).
\end{IEEEbiography}

\vspace{-80 mm}
\begin{IEEEbiography}[{\includegraphics[width=1in,height=1.0in,clip,keepaspectratio]{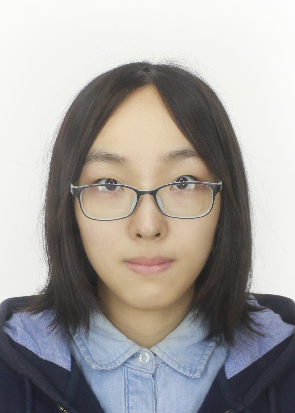}}]{Ruichen Zuo} is a undergraduate student in Beijing Institute of Technology. Her major is electronic information engineering and she is also interested in brain-inspired Spiking Neural Networks.
\end{IEEEbiography}


\vspace{-80 mm}
\begin{IEEEbiography}[{\includegraphics[width=1in,height=1.0in,clip,keepaspectratio]{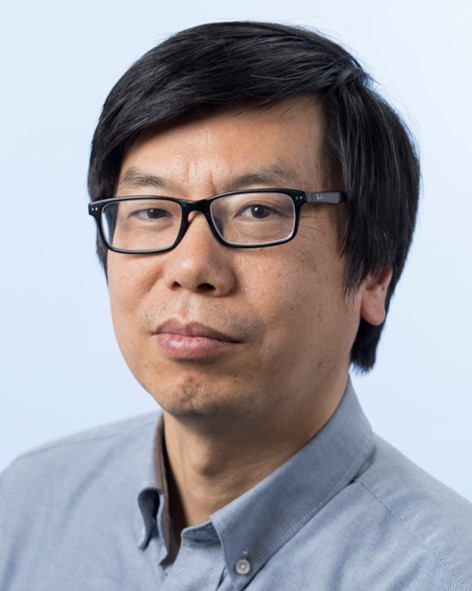}}]{Bo Xu}
is a professor, the director of the Institute of Automation Chinese Academy of Sciences, and also deputy director of the Center for Excellence in Brain Science and Intelligence Technology, Chinese Academy of Sciences. His main research interests include brain-inspired intelligence, brain-inspired cognitive models, natural language processing and understanding, brain-inspired robotics.
\end{IEEEbiography}


\end{document}